\documentclass{article}

\PassOptionsToPackage{numbers, compress}{natbib}

 \usepackage[final]{neurips_2025}

\usepackage[utf8]{inputenc} 
\usepackage[T1]{fontenc}    
\usepackage{hyperref}       
\usepackage{url}            
\usepackage{booktabs}       
\usepackage{amsfonts}       
\usepackage{nicefrac}       
\usepackage{microtype}      
\usepackage{xcolor}         
\usepackage{graphicx}       
\usepackage[frozencache,cachedir=minted-cache]{minted}
\usepackage[most]{tcolorbox}
\tcbuselibrary{breakable,skins}
\usepackage{dblfloatfix}      

\newcounter{supprompt}
\renewcommand{\thesupprompt}{P\arabic{supprompt}}

\newcommand{\PromptTitle}[1]{%
  \textbf{\thesupprompt:}~#1%
  \ifx\PromptMetaDefault\empty\else\\[1pt]\footnotesize \PromptMetaDefault\fi
}
\newcommand{\PromptMetaDefault}{} 
\newenvironment{SuppPromptWide}[2][]{%
  \refstepcounter{supprompt}%
  \begin{tcolorbox}[
    enhanced, breakable,
    colback=black!1, colframe=black!35, boxrule=0.4pt,
    title={\PromptTitle{#2}},
    left=1mm, right=1mm, top=1mm, bottom=1mm,
    width=\linewidth 
  ]%
  \if\relax\detokenize{#1}\relax\else\phantomsection\label{#1}\fi%
}{%
  \end{tcolorbox}%
}
\makeatletter
\renewcommand{\@noticestring}{%
{Accepted at ML4Molecules 2025} (ELLIS UnConference workshop), Copenhagen, Denmark, December 2, 2025.%
}
\makeatother

\title{Grounding Large Language Models in Reaction Knowledge Graphs for Synthesis Retrieval}

\author{%
  \parbox{\textwidth}{\centering
    \textbf{Olga Bunkova\textsuperscript{1}} \quad 
    \textbf{Lorenzo Di Fruscia\textsuperscript{1}} \quad 
    \textbf{Sophia Rupprecht\textsuperscript{2}} \quad 
    \textbf{Artur M. Schweidtmann\textsuperscript{2}} \\ [12pt]
    \textbf{Marcel J.T. Reinders\textsuperscript{1}} \quad 
    \textbf{Jana M. Weber\textsuperscript{1,*}} \\[12pt]
    \normalfont
    \textsuperscript{1}Department of Intelligent Systems, Delft University of Technology, The Netherlands \\
    \textsuperscript{2}Department of Chemical Engineering, Delft University of Technology, The Netherlands \\
    {\small \texttt{\{l.difruscia, s.rupprecht, a.schweidtmann, m.j.t.reinders, j.m.weber\}@tudelft.nl}} \\
    \small *Corresponding Author
  }%
}

\begin{document}

\maketitle

\begin{abstract}
Large Language Models (LLMs) can aid synthesis planning in chemistry, but standard prompting methods often yield hallucinated or outdated suggestions. We study LLM interactions with a reaction knowledge graph by casting reaction path retrieval as a Text2Cypher (natural language to graph query) generation problem, and define single- and multi-step retrieval tasks. We compare zero-shot prompting to one-shot variants using static, random, and embedding-based exemplar selection, and assess a checklist-driven validator/corrector loop. To evaluate our framework, we consider query validity and retrieval accuracy. We find that one-shot prompting with aligned exemplars consistently performs best. Our checklist-style self-correction loop mainly improves executability in zero-shot settings and offers limited additional retrieval gains once a good exemplar is present. We provide a reproducible Text2Cypher evaluation setup to facilitate further work on KG-grounded LLMs for synthesis planning. \textcolor{black}{Code is available at \url{https://github.com/Intelligent-molecular-systems/KG-LLM-Synthesis-Retrieval}}.
\end{abstract}

\section{Introduction}

Large Language Models (LLMs) have transformed Natural Language Processing (NLP) by enabling tasks such as question answering and text summarization through large-scale pretraining on text corpora \cite{Minaee2024LargeSurvey}. A key capability of LLMs is In-Context Learning (ICL), where models perform new tasks solely conditioned on instructions or demonstrations with no parameter updates \cite{brownLanguageModelsAre2020}. Prompt engineering, the systematic design of inputs to improve outputs, has become essential to harness ICL effectively. Beyond language, LLMs are increasingly applied in various specialized fields, one of them being cheminformatics. The reason for this is their ability to process and interpret molecular data in string format \cite{Zhang2024ScientificDomains}.

Chemical synthesis or pathway planning seeks to design a sequence of reactions that produce a target molecule from available precursors or vice versa \cite{Strieth-Kalthoff2024ArtificialKnowledge, Gricourt2024ArtificialReview}. With given data, the task requires the identification of optimal sequences of reactions, both forward towards products and backwards towards precursors \cite{weberChemicalDataIntelligence2021, vollReactionNetworkFlux2012, weberDiscoveringCircularProcess2022}. In the context of synthesis in unknown chemical space, the task may involve forward prediction, where products are generated from reactants, or retrosynthetic predictions, where candidate precursors for a desired molecule are predicted \cite{schwallerMolecularTransformerModel2019, kreutterPredictingEnzymaticReactions2021, probstBiocatalysedSynthesisPlanning2022}\textcolor{black}{ \cite{seidlImprovingFewZeroShot2022}}. In both cases, extending this to multi-step planning requires chaining single-step retrievals/predictions into coherent routes, typically through search algorithms \cite{Gricourt2024ArtificialReview, Liu2024EvaluatingPrediction} \textcolor{black}{\cite{seglerPlanningChemicalSyntheses2018}}.

LLMs offer promising reasoning abilities for reaction planning. Among others, they could function as flexible conversation agents to help assemble all possible synthesis routes, suggest alternative/optimized routes, or help evaluate suggested pathways. Yet they remain prone to hallucinations and outdated knowledge in fast-moving chemical domains \cite{Tonmoy2024AModels}, \cite{Minaee2024LargeSurvey}. Retrieval Augmented Generation (RAG) \cite{Lewis2020Retrieval-AugmentedTasks} partially mitigates these issues, but generally relies on isolated retrieval and not a single, connected view of molecules and reactions across sources. Knowledge Graphs (KGs) provide a more structured alternative by encoding molecules and reactions as entities and relations, preserving both short (direct precursors/products) and long-range links (multi-step synthesis routes) \cite{Perozzi2024LetLLMs}, \cite{Edge2024FromSummarization}. This representation enables path-constrained retrieval and multi-hop reasoning \cite{Kau2024CombiningModels,Ibrahim2024AChallenges}, offering a foundation for grounding LLM predictions in verified reaction data. A practical way to enable this interaction is to pose questions as declarative graph queries in \textit{Cypher} language \cite{francisCypherEvolvingQuery2018}.

\textcolor{black}{While Text2Cypher has been explored for general KGs \cite{Ozsoy2024Text2Cypher:Databases, hornsteinerRealTimeTexttoCypherQuery2024, agrawalCanKnowledgeGraphs2024a}) there is, to our knowledge,} no prior chemistry-specific, execution-grounded evaluation of Text2Cypher tasks (i.e., generating Cypher queries from natural language instruction) on reaction KGs. In this study we assess whether LLMs can generate executable Cypher queries that yield correct single- and multi-step retrosynthesis retrieval over a bipartite KG. \textcolor{black}{Our findings translate into operating guidance: We contribute an evaluation protocol, metrics, and design recommendations for reliable Text2Cypher retrieval over reaction KGs. Furthermore, }we analyze how demonstration choice and a lightweight correction loop affect the executed results across prompting strategies and tasks.

\section{Methods}

\begin{figure}[h!]
    \centering
    \includegraphics[width=1.\linewidth]{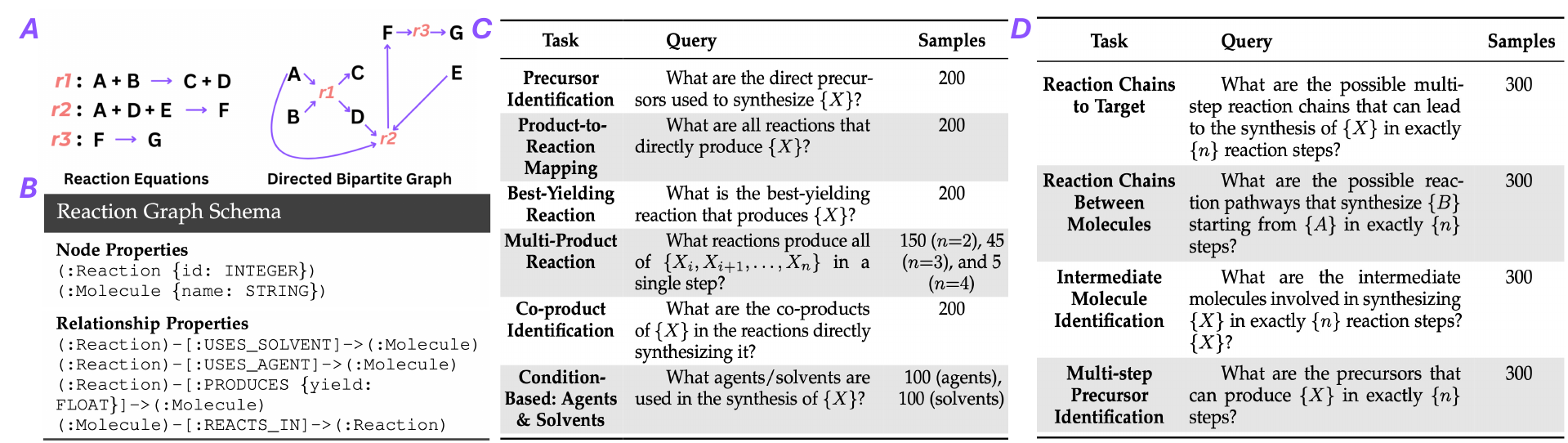}
    \caption{(A) Chemical reaction equation with reactants and products (left), and a possible directed bipartite graph representation, where nodes correspond to reactions and molecules (right). (B) Schema used for the reaction knowledge graph. (C, D) Retrosynthesis task types for single- and multi-step reaction retrieval, along with natural language queries and number of task samples. For multi-step reaction retrieval $n \in\{2,3,4\}$.}
    \label{fig: method_1}
\end{figure}

\paragraph{Reaction Knowledge Graph: Data and construction}
We build our dataset using USPTO reactions in SMILES format \cite{Wigh2024ORDerly:Data, weiningerSMILESChemicalLanguage1988a}, removing duplicates, canonicalizing SMILES strings, and filtering out rare entries. We retain reactions with at most four reactants, four products, four agents, and four solvents, which cover $\sim$95\% of the entire dataset.\newline
Individual chemical reactions can naturally be modeled using different graph-based representations. We rely on a Bipartite Graph (BG) for the KG construction. A BG contains two types of nodes connected only across types, as shown in Figure \ref{fig: method_1}A. Each individual reaction is represented as a bipartite graph with two distinct node types: \textbf{(:Reaction)}, identified by a unique \texttt{id}, and \textbf{(:Molecule)}, uniquely identified by a canonicalized SMILES \texttt{name}. This representation preserves the multi-component context of a reaction and the required \texttt{Mol$\rightarrow$ Reac$\rightarrow$ Mol} directionality for path reasoning \cite{Klamt2009HypergraphsNetworks,Muller2022Whatchemical,Garcia-Chung2024ChemicallyHypergraphs}. The full reaction schema is shown in Figure \ref{fig: method_1}B. For efficiency of this study, we randomly sample 50k reactions and load them into \texttt{Neo4j} \cite{Neo4j}.

\paragraph{Tasks: Single- and Multi-step retrieval}
We evaluate our framework on two distinct types of tasks. Single-step tasks target one-hop reaction context (reactants, products, agents/solvents). They focus on different aspects of retrosynthesis and aim to evaluate the ability of the LLM to generate valid Cypher queries of varying syntactic complexity in a chemically constrained scenario. Multi-step tasks target paths of length \(\le L\) linking precursors to a product (\(L\in\{2,3,4\}\)) under the bipartite directionality, representing a broader chemical scenario. The objective for single-step synthesis always requires retrieving the full reaction context (reactants, products, solvents etc.), whereas for multi-step synthesis, it requires retrieving ordered sequences of reaction nodes.\newline
We randomly sample molecules from the KG, and instantiate queries by pairing natural language question templates with a reference Cypher template (sequential \texttt{MATCH}/\texttt{OPTIONAL MATCH} with \texttt{RETURN}/\texttt{COLLECT}) and the corresponding gold, i.e. correct, answers. The single-step setting comprises 1200 queries over six tasks (200 per type). The multi-step setting comprises 1200 queries over four tasks (300 per type), ensuring reachable paths of the specified length. We report our query categories in Tables \ref{fig: method_1}C and \ref{fig: method_1}D.

\paragraph{Text2Cypher prompting and exemplar selection}
We investigate how our prompting choices affect retrieval over the KG. We design five prompt versions per setting (single- ~\ref{s-p1}-~\ref{s-p5}/multi-step ~\ref{m-p1}-~\ref{m-p5}), increasingly adding instructions/context from Prompt one to five. Additionally, we compare zero-shot (\textit{ZS}) prompting to various one-shot prompting strategies: Static (\textit{1S}, one exemplar is selected and reused for all tasks), dynamic random (\textit{1S-D-R}, for each query one exemplar is randomly selected from a predefined example bank), and dynamic semantic (\textit{1S-D-S}, an exemplar is retrieved from a predefined example bank stored in a vector database, based on vector similarity) schemes. We illustrate the pipeline schema in Figure \ref{fig:method_2}A. More details on the prompting strategies are reported in the Appendix section \ref{App: prompt}.

\paragraph{Checklist validator/corrector}
We implement a Chain-of-Verification (CoVe) style loop to address common generation errors. After query generation from natural language, a checklist, consisting of the most frequent error types from prior error analysis, validates the executability of the query in Neo4j. On failure, the LLM receives the error message and proposes a corrected query; we allow up to three attempts before termination. We illustrate the CoVe pipeline in Figure \ref{fig:method_2}B.

\paragraph{Framework evaluation and implementation}
We evaluate our framework based on two concepts: Query evaluation and retrieval evaluation.
For query evaluation, we compute text-to-text similarity with the metrics BLEU, METEOR, and ROUGE-L \cite{Ozsoy2025EnhancingFiltering} to compare the generated and reference Cypher queries. These metrics vary in their strategy for comparing text sequences, and are further defined in the Appendix section \ref{App: subsec evaluation}. \newline
For retrieval evaluation, retrieved results are collected as lists of dictionaries and compared to gold answers with micro-averaged precision, recall, and F1 scores across relevant keys (\texttt{reactants}, \texttt{agents}, \texttt{solvents} etc.). We show our scheme of the gold answer format for both in Figure \ref{fig:method_2}C. For multi-step routes, we report exact-path precision, recall, and F1 scores (treating each path as an ordered set of reaction node ids) and Partial Path Recall (PPR), which gives partial credit when a predicted path matches a continuous terminal fragment of a gold path. \newline
Experiments use \texttt{GPT-4.1-mini-2025-04-14} \cite{openai2025gpt41mini} via the OpenAI API with \(T{=} 0\) to ensure deterministic decoding. Additional implementation details are provided in Appendix Section \ref{App: implementation}.

\begin{figure}[h!]
    \centering
    \includegraphics[width=.9\linewidth]{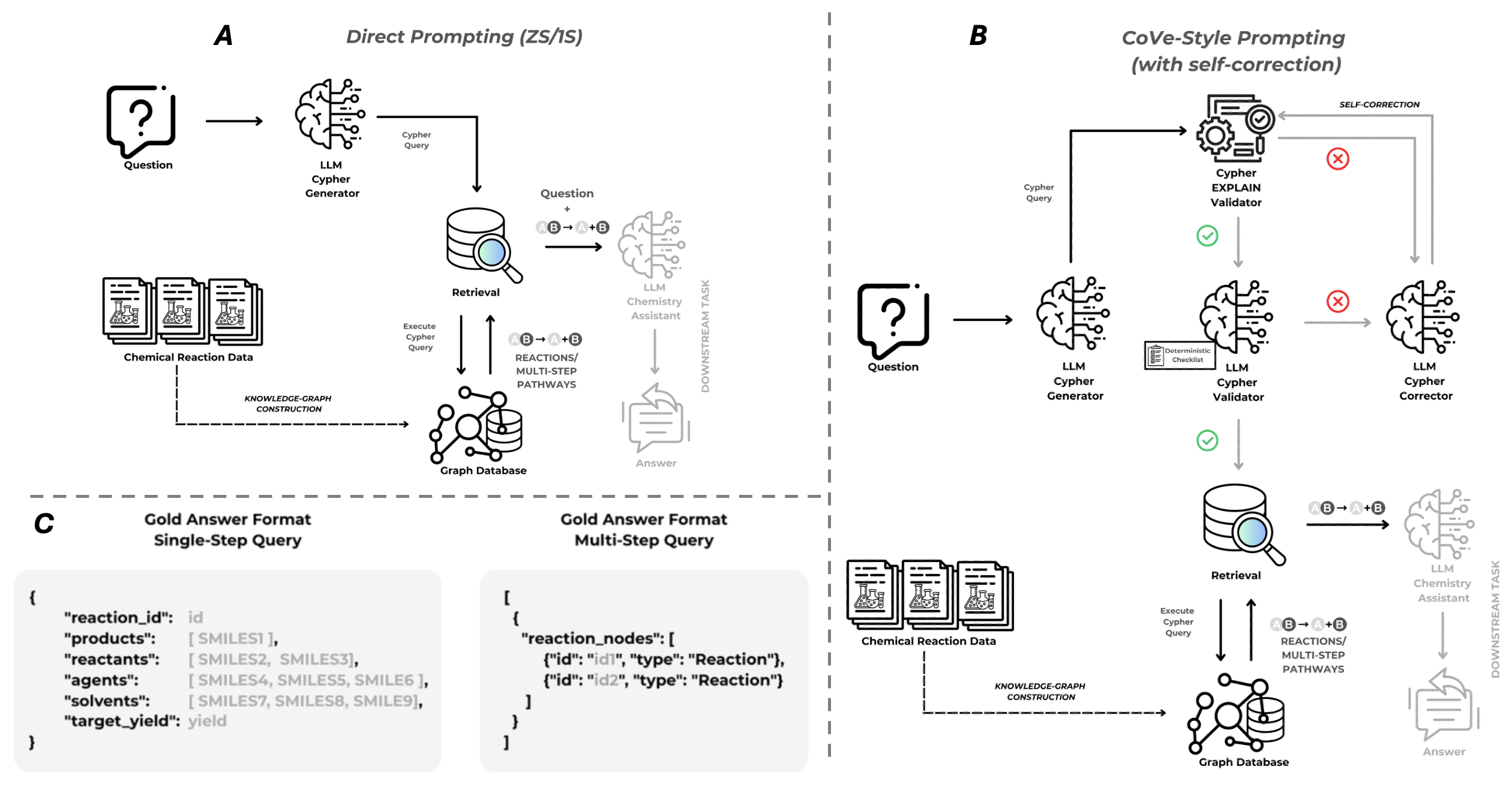}
    \caption{(A) Based on the user question, with (one-shot) or without (zero-shot) an exemplar, an LLM generates a Cypher query with the goal of retrieving relevant reactions/pathways. The query is then executed on the BG in Neo4j. (B) CoVE-style prompting. First, an LLM generates a candidate query. A validator LLM checks it against a fixed checklist, and outputs a list of specific errors if applicable. A corrector LLM then applies minimal edits only to resolve the flagged issues. The process repeats until either the query becomes valid or a maximum of three correction attempts are reached. (C) Format of gold answers for single-step and multi-step retrieval tasks.}
    \label{fig:method_2}
\end{figure}

\section{Results}

Our framework reliably retrieves single- and multi-step reaction information from the KG given natural language questions across most tasks. In the following, we examine which evaluation criteria are most informative and which design choices most affect retrieval quality. 

\begin{figure}[h!]
    \centering
    \includegraphics[width=.95\linewidth]{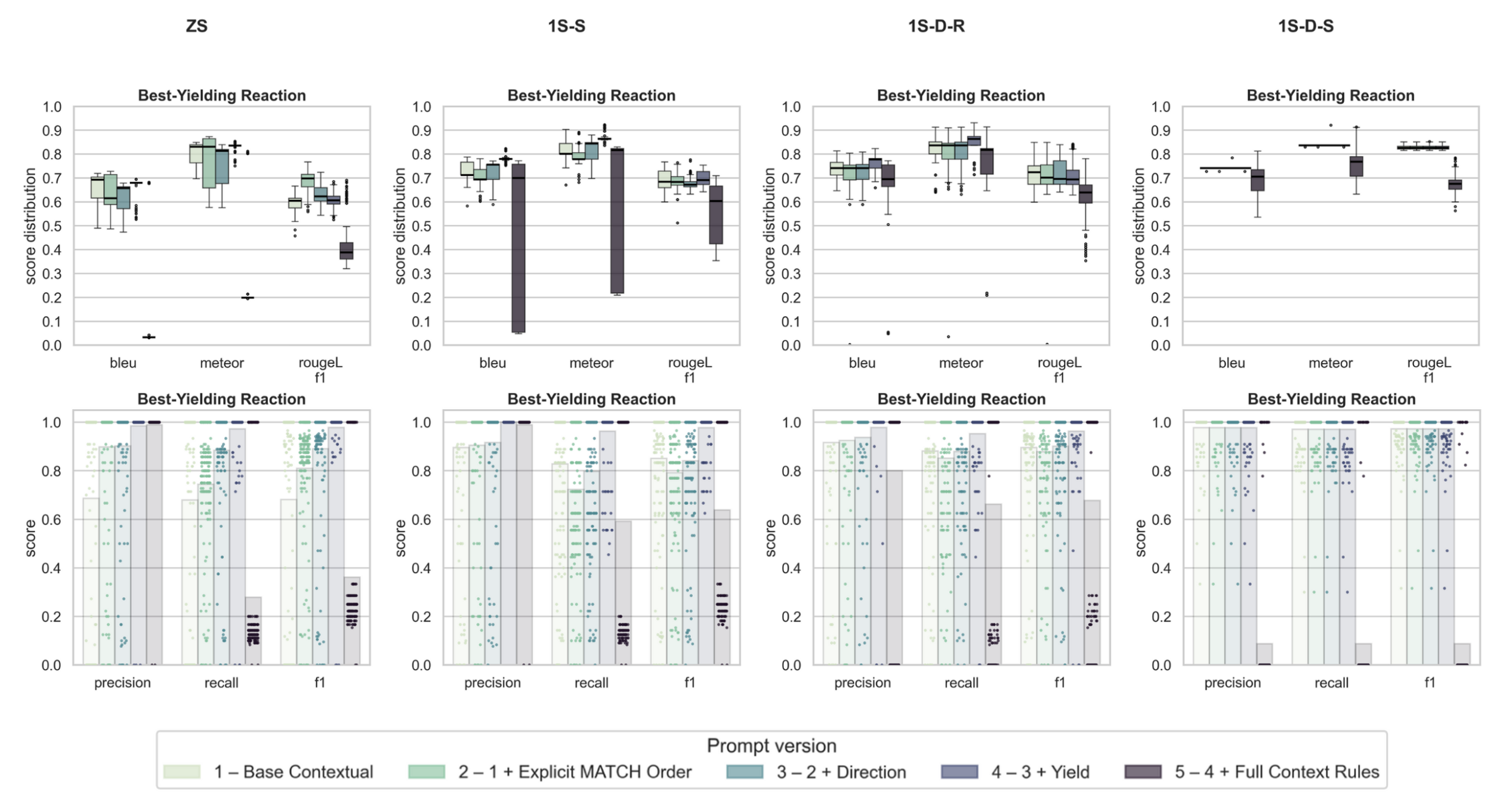}
    \caption{Text-to-text similarity (for Cypher queries, top row) and retrieval metrics (bottom row) for the single-step task \texttt{Best-Yielding Reaction} under different prompt settings: Zero-shot (ZS), one-shot static (1S-S), one-shot random (1S-D-R), and one-shot semantic (1S-D-S). Within each setting, five prompt versions (indicated in the bottom legend) add increasing contextual/structural guidance.}
    \label{fig:ST_plot}
\end{figure}

\paragraph{Text-to-text similarity is a poor proxy for retrieval accuracy}
We observe that text-to-text similarity between queries measured by BLEU, METEOR, and ROUGE-L does not represent a good indicator of retrieval success. Two mechanisms explain the divergence. First, multiple Cypher formulations can be semantically equivalent and return the same answers despite different structures. Conversely, structurally similar queries with small syntactic shifts (e.g., losing edge directionality) keep the query overlap high while breaking or altering the retrieval. This mismatch can be observed in Figure \ref{fig:ST_plot}, where retrieval F1 score can vary a lot (e.g. one-shot random vs one-shot semantic, Prompt one to three), with comparable text-to-text similarities for generated Cypher queries. Extended results for single-step tasks are illustrated in Appendix Figures \ref{App: fig: ST query}, \ref{App: fig: ST retrieval} and \ref{App: tab: surface equiv}.

\begin{figure}[h!]
    \centering
    \includegraphics[width=1.\linewidth]{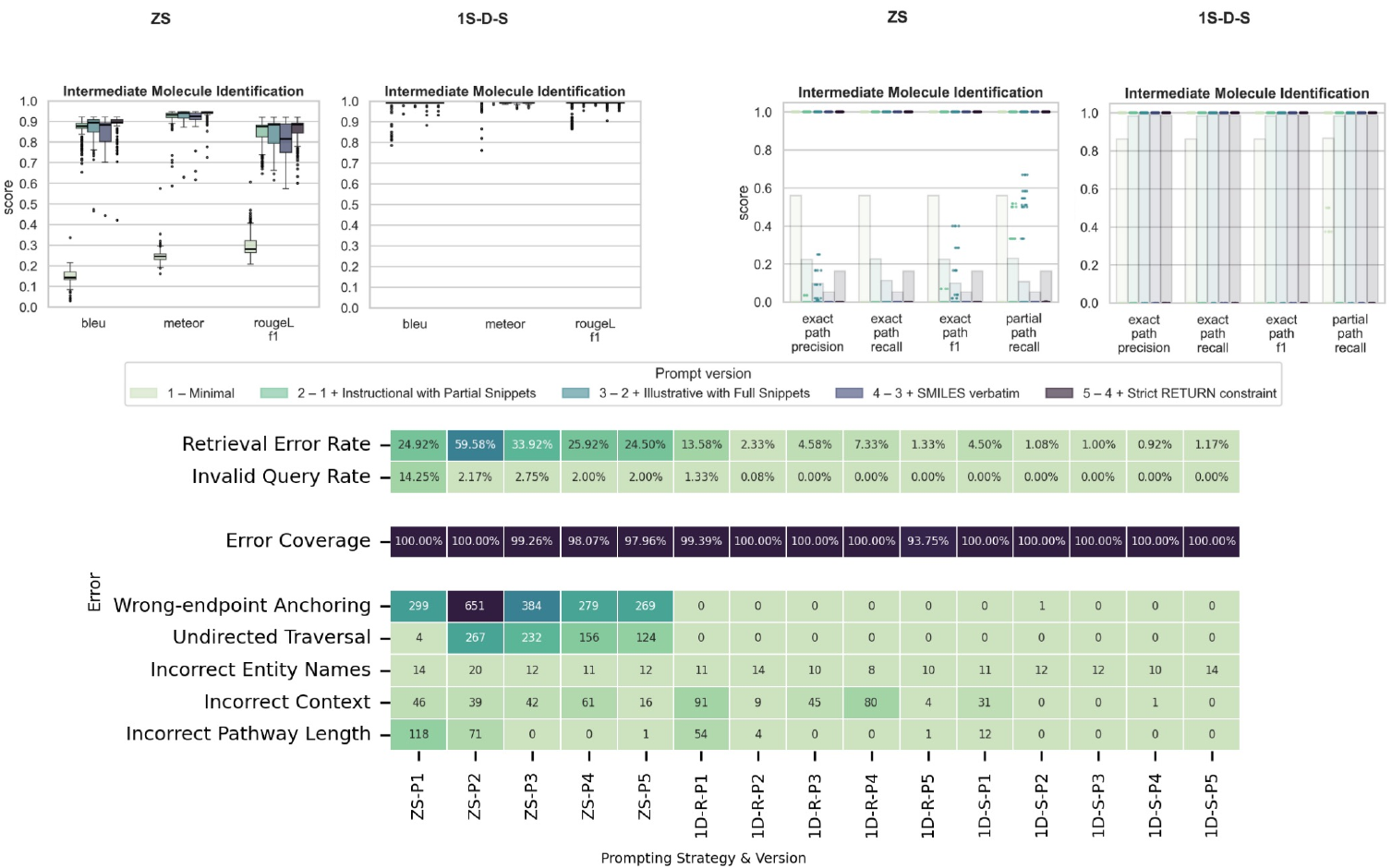}
    \caption{Text-to-text similarity (for Cypher queries, top, left) and retrieval metrics (top, right) for the multi-step task \texttt{\small{Intermediate Molecule Identification}} under zero-shot (ZS) and one-shot semantic (1S-D-S). One-shot largely removes zero-shot failures (anchoring, traversal direction, pathway length). Within each setting, five prompt versions add increasing contextual/structural guidance. The error taxonomy table (bottom) reports the rate of non-executable queries, incorrect retrievals, and their absolute counts across five error categories for all multi-step tasks, which cover almost all observed cases.}
    \label{fig:MT_plot}
\end{figure}

\paragraph{Largest performance gains from zero- to one-shot in multi-step tasks}
Moving from zero- to one-shot prompting removes most common retrieval errors. 
In multi-step task queries, the most occurring failures are \textit{endpoint anchoring} (treating the target molecule as the start of the reaction pathway) and traversal-direction violation of the bipartite path. These errors are largely removed when moving from zero- to one-shot prompting, regardless of the tested strategy: Careful selection of a good exemplar instead of a random one, or choosing a different prompt version, offer only marginal gains. We show a representative task in Figure \ref{fig:MT_plot}, where the error taxonomy table reports that \textit{invalid query} and \textit{retrieval error} rates decrease when an example is provided in the input Prompt. Extended results are illustrated in Appendix Figures \ref{App: fig: MT query} and \ref{App: fig: MT retrieval}. 

\paragraph{Checklist self-correction (CoVe) has limited benefits, validator is the bottleneck}
The CoVe-loop mainly assists in the zero-shot setting: It mostly repairs generated queries by reducing completeness errors, i.e. missing reaction components (reactants, products, agents). For example, \textit{Reactants missing} and \textit{Products missing} retrieval errors go down by $\sim 80\%$ in absolute count for single-step tasks in Prompt 2 in a zero-shot setting when CoVe is used, while in this case we see no clear improvement using the CoVe in the one-shot semantic settings. We report a summary of the retrieval errors in the Appendix Table \ref{App:Tab: cove}. The limiting factor is the validator, not the correction step: A generic checklist misses up to $\sim$ 86-95\% (\textit{Non-detected error rate}) of task-specific failures such as duplicate or incomplete molecules. We recommend investing in task-specific/schema-aware validators for future CoVe-loops.


\section{Conclusions}

We present a Text2Cypher pipeline over a reaction KG for chemical pathway planning, converting natural language questions into graph queries for both single- and multi-step reactions.
We find that execution-grounded evaluation is essential: Text-to-text similarity for queries can be high while retrieval is incorrect. We further show that largest performance gains arise when moving from zero- to one-shot prompting, especially in the multi-step setting, with best results when the exemplar query is similar to the target query (task/intent). Lastly, a checklist-driven self-correction loop primarily improves executability on weak baselines, and yields only modest retrieval gains if an exemplar is already provided.
Our results hold for the selected LLM and the designed knowledge graph, templates, and validator checks. Broader model comparisons, larger KGs, and task-specific validators are natural next steps. Our results provide practical guidelines for KG-grounded retrieval for reaction planning and indicate both challenges and exciting promises towards LLM-based reaction planning workflows.

\clearpage
\bibliographystyle{plainnat}
\bibliography{references}   

\newpage
\section{Appendix}
\setcounter{table}{0}
\renewcommand{\thetable}{A\arabic{table}}
\setcounter{figure}{0}
\renewcommand{\thefigure}{A\arabic{figure}}

\subsection{Prompt strategies}
\label{App: prompt}
In our setting, prompt engineering starts with a baseline, minimal prompt. Based on subsequent error analysis, additional instructions and constraints are incrementally introduced to further improve the model’s context: It includes relevant schema and formatting rules, dataset-specific nuances, and guidance on the retrieval of full reaction contexts. The user prompt contains the input question, whereas the system prompt defines the model’s role as a helpful assistant for generating Cypher queries from natural language.
In total, 5 Prompts were tested during the prompt engineering experiments (see Prompts ~\ref{s-p1}-~\ref{s-p5}) for the single-step tasks, and 5 Prompts for the multi-step task (see Prompts ~\ref{m-p1}-~\ref{m-p5}), where each prompt was designed following OpenAI’s guidelines on prompt engineering \cite{OpenAIGuide}. 

\paragraph{Static one-shot prompting}
\label{App: static one-shot}
In this setting, a single representative example is selected for all tasks. In particular we select an example corresponding to the most common query pattern. For the single-step example bank, it is $\small{\texttt{Product Identification}}$ (shared across three out of six query types). For the multi-step example bank, these are $\small{\texttt{Multi-Step Product Discovery}}$ and $\small{\texttt{Forward Synthesis Intermediate Identification}}$. The selected examples are kept the same across all tasks.

\paragraph{Dynamic one-shot: Random selection}
In this setting, we randomly select an example from the corresponding example bank. Since the chosen example may be dissimilar from the given input question, it serves as a baseline for evaluating exemplar quality for Cypher generation.

\paragraph{Dynamic one-shot: Embedding-based selection}
In this setting, the example bank is stored in a vector database. We use the sequence model listed in \ref{App: implementation} to compute the embeddings for each example. The input question is embedded in the same way, and we select the one-shot exemplar by top-1 cosine similarity. To prevent chemistry specific influence, SMILES strings are masked in all queries, so similarity reflects task intent rather than chemical content. Because the example bank contains "opposite" tasks (forward vs retrosynthesis), logical intent descriptions are also added to all demonstrations, to prioritize \textit{structural alignment} on the underlying task over surface-level semantic similarity. Here structural alignment denotes exemplar/query pairs sharing the same Cypher pattern type.

\subsection{Exemplar selection for dynamic one-shot}

\begin{figure}[h!]
    \centering
    \includegraphics[width=1.\linewidth]{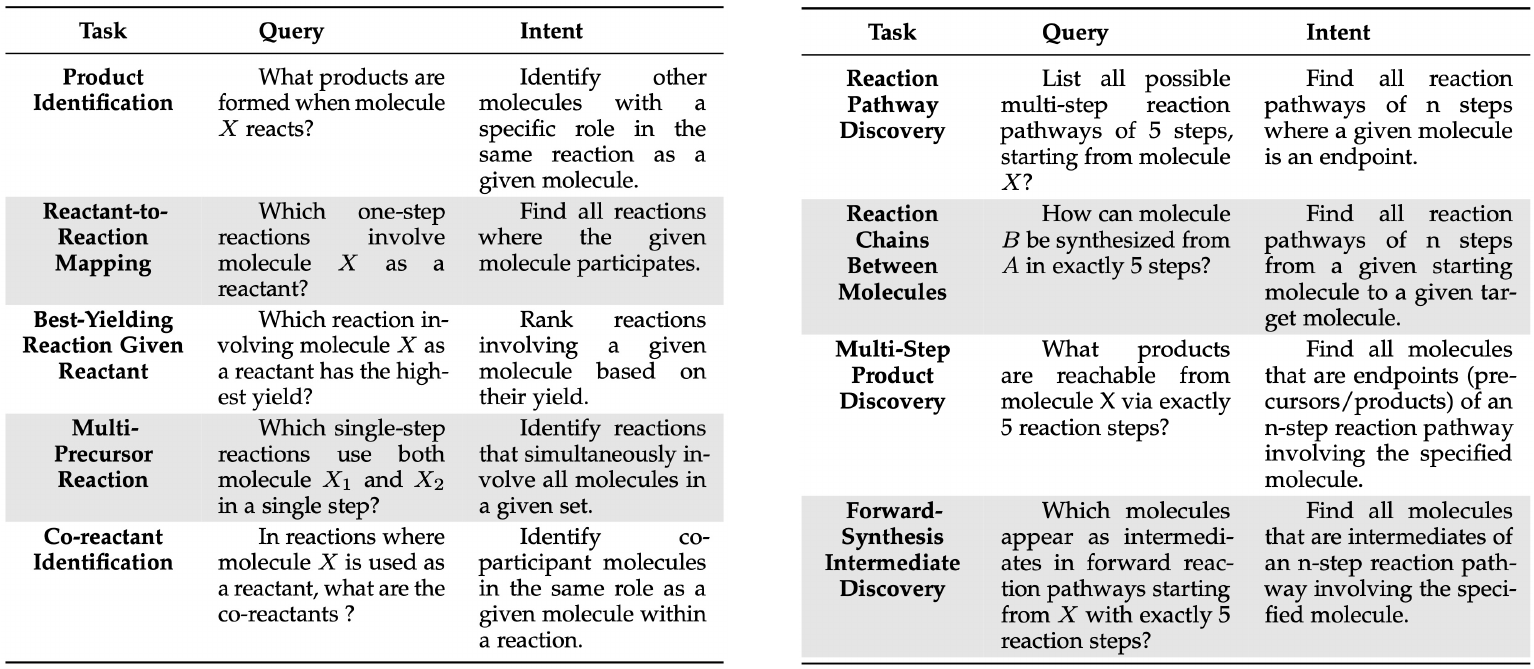}
    \caption{Example banks for single-step (left) and multi-step (right) retrieval. Alongside each task we report their natural language queries and corresponding logical intent.}
    \label{App:Tab data bank}
\end{figure}

In one-shot prompting strategies, the system prompt remains unchanged, containing the same guidelines as in a
zero-shot setting, with an example pair of natural language question and the corresponding Cypher query now included in the user prompt. Exemplars for the one-shot setting are always extracted from an example bank. It contains questions similar in intent to those in the test dataset, but formulated in the reverse direction as forward synthesis tasks. This prevents the LLM from ``copying” query patterns directly and instead evaluates how well it generalizes on the Text2Cypher task. The example banks are different from single- and multi-step tasks. We summarize them in Tables \ref{App:Tab data bank}. Note that SMILES strings are intentionally excluded from the examples to prevent biasing Cypher generation with chemical context.

\subsection{Chain-of-Verification (CoVe)}

The Chain-of-Verification method is a self-criticism framework designed to verify LLM-generated responses. Given a generated answer, the model also creates a checklist to verify it. After each point on the checklist is addressed and the context is aggregated, a refined response is produced \cite{Schulhoff2024TheTechniques}. After the LLM generates the candidate query, its executability is checked with \texttt{EXPLAIN} in Neo4j. If the query is executable, a deterministic directionality corrector is applied. If the query is not executable, an LLM corrector is called to fix it based on the error message. For that, all SMILES strings are masked with placeholders to avoid special-character
interference. The next step is the LLM validation against a fixed, task-specific checklist derived from prior error analysis. If the query is classified as valid, it is executed; otherwise, it is passed again to the LLM corrector to apply minimal edits based on the identified issues. The process repeats until either the validator labels the query as correct or a limit of three correction attempts is reached. \newline

\begin{figure}[h!]
    \centering
    \includegraphics[width=1.4\linewidth, angle=-90]{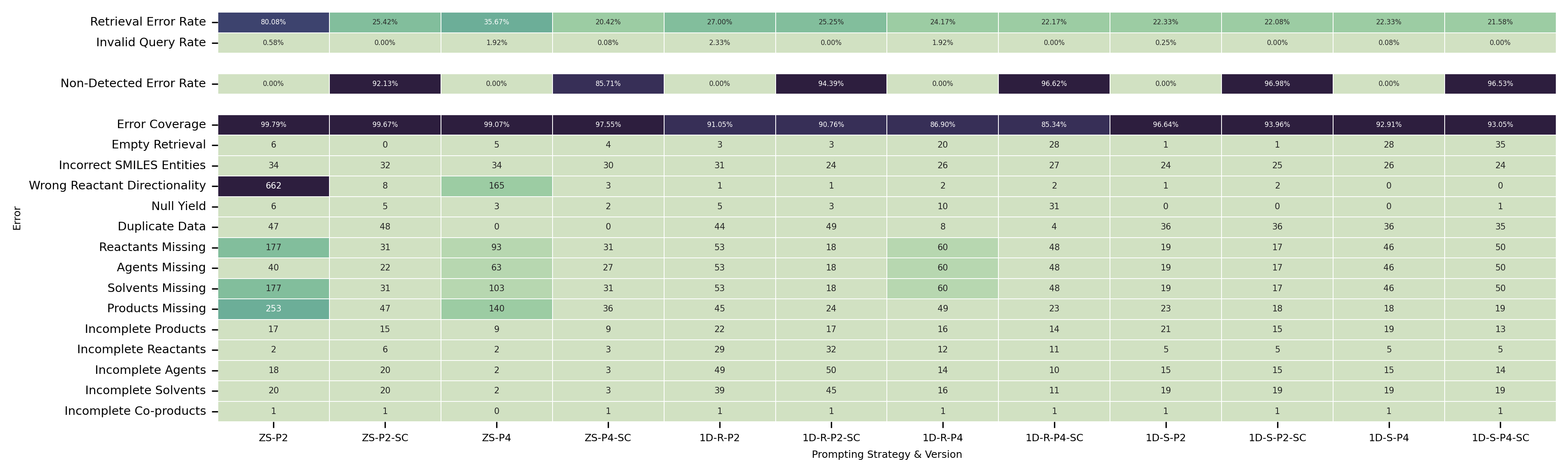}
    \caption{Distribution of retrieval errors for direct prompting vs. checklist-driven (CoVe style) prompting for single-step tasks. Error analysis covers three prompting strategies: Zero-shot (ZS), one-shot random (1S-D-R), and one-shot semantic (1S-D-S) and two Prompt versions, P2 and P4, with and without self-correction. \textit{Retrieval error rate} is the fraction of executable queries with wrong retrieved context. \textit{Non-detected error rate} is the fraction of retrieval error cases not flagged and corrected by the CoVe pipeline. \textit{Error coverage} is the fraction of total retrieval errors that are assigned to the specific categories shown below.}
    \label{App:Tab: cove}
\end{figure}

We report the distribution of retrieval errors for two Prompt version, P2 and P4, with and without self-correction, in Figure \ref{App:Tab: cove}. 
\textit{Wrong Reactant Directionality} errors are corrected by the deterministic directionality corrector, not by the LLM corrector, which explains most of the improvement for the zero-shot setting in the retrieval rate.

\subsection{Implementation details}
\label{App: implementation}
The USPTO dataset was preprocessed using the ORDerly \cite{ORDerly} library, to reduce inconsistencies and improve data quality.
For demonstration selection we use the Sentence Transformer model \texttt{all-mpnet-base-v2} \cite{sentence-transformers-all-mpnet-base-v2}. The sentence encoder is \texttt{BiomedNLP-BiomedBERT}\cite{microsoft-biomedbert-abstract} (also available via \texttt{Hugging Face}), and the similarity threshold for cosine similarity is $0.93$.
Orchestration uses \texttt{LangChain} for prompting and \texttt{LangGraph} for workflow control; the KG backend is \texttt{Neo4j}.


\subsection{Evaluation}
\label{App: subsec evaluation}
\paragraph{Surface metrics for generated Cypher}

The generated Cypher queries are compared to the reference queries using BLEU, METEOR and ROUGE-L scores. Since Text2Cypher is a text-to-text generation task, these metrics are used for evaluating their similarity:
\begin{itemize}
    \item \textbf{BLEU}: It measures how many n-grams (contiguous word sequences) in the generated text also appear in the reference text. It focuses on exact word overlap and is precision-oriented;
    \item \textbf{METEOR}: Extends BLEU by accounting for stemming (reducing words to their root form), synonyms, and word order, combining both precision and recall into a single score. It better captures meaning similarity than exact wording;
    \item \textbf{ROUGE-L}: Evaluates the longest common subsequence between the generated and reference texts, emphasizing how well the overall sequence and structure of words are preserved.
\end{itemize}
They all range from 0 to 1, where higher values indicate greater fidelity to the reference text.

\paragraph{Retrieval metrics}

\begin{figure}[h!]
    \centering
    \includegraphics[width=1\linewidth]{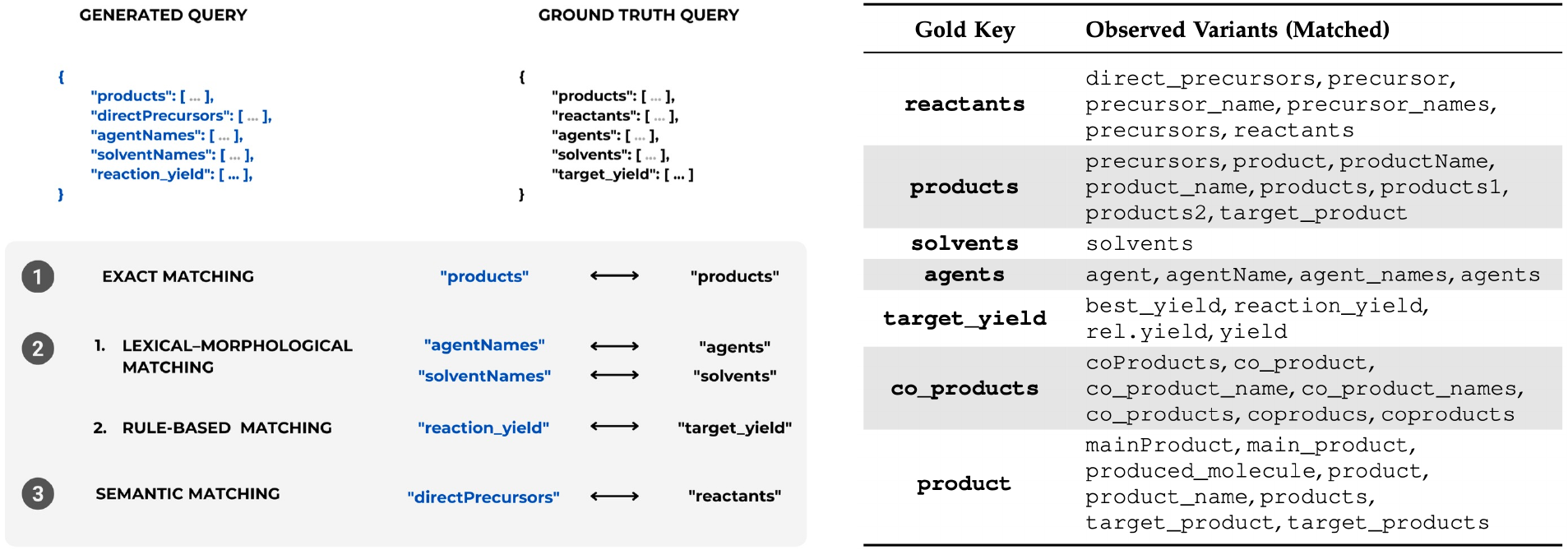}
    \caption{Key-matching pipeline for single-step tasks (left), from generated query to ground truth, through sequential matching steps: Exact$\rightarrow$ lexical$\rightarrow$ semantic. On the right, a table of key variants observed and matched during the procedure.}
    \label{App: Fig: key-matching}
\end{figure}

The retrieved results are represented as lists of dictionaries with specific keys. Generated queries can present different dictionary structures and different key names compared to the reference query. For this reason, a key-matching procedure is required. We first normalize each key by lowercasing, trimming whitespace, converting camelCase to snake\_case, and removing trailing \textit{name}/\textit{names}. The matching process involves three stages, illustrated in Figure \ref{App: Fig: key-matching}:
\begin{itemize}
    \item \textbf{Exact matching}: Normalized predicted keys are matched to ground truth keys based on string equality;
    \item \textbf{Rule-based and lexical matching}: Stemming is performed to reduce keys to their root form (e.g., agents$\rightarrow$ agent);
\item \textbf{Embedding-based matching (semantic)}: Remaining unmatched predicted keys are compared to gold keys using embedding similarity. Each prediction is assigned to the highest scoring key if the score $~\ge$ \textit{threshold}. Notably, multiple predicted keys may map to a single gold key, so we aggregate their values during evaluation.
\end{itemize}

In the single-step setting, for each sample, True Positives, False Positives, and False Negatives are computed. They are then used to compute Precision, Recall and F1 metrics. These metrics are computed separately for each key (e.g. \textit{products}, \textit{reactants}, \textit{solvents}), before sample-aggregation, so we refer to them as micro-averaged.

In the multi-step setting, the gold answer is a list of distinct reaction paths, represented by an ordered sequence of reaction nodes. Two paths are considered identical if their ordered \texttt{id} sequences are the same, eliminating the need for key matching. The two main metrics here are Exact Path Retrieval, that assigns a score of 0 to any incorrect path and 1 to exact path overlaps, and Partial Path Recall, which measures how much of the retrieved route overlaps with a continuous subpath of the reference (gold) pathway.

\subsection{Single- and multi-step task results}

\begin{figure}[h!]
    \centering
    \includegraphics[width=1.\linewidth]{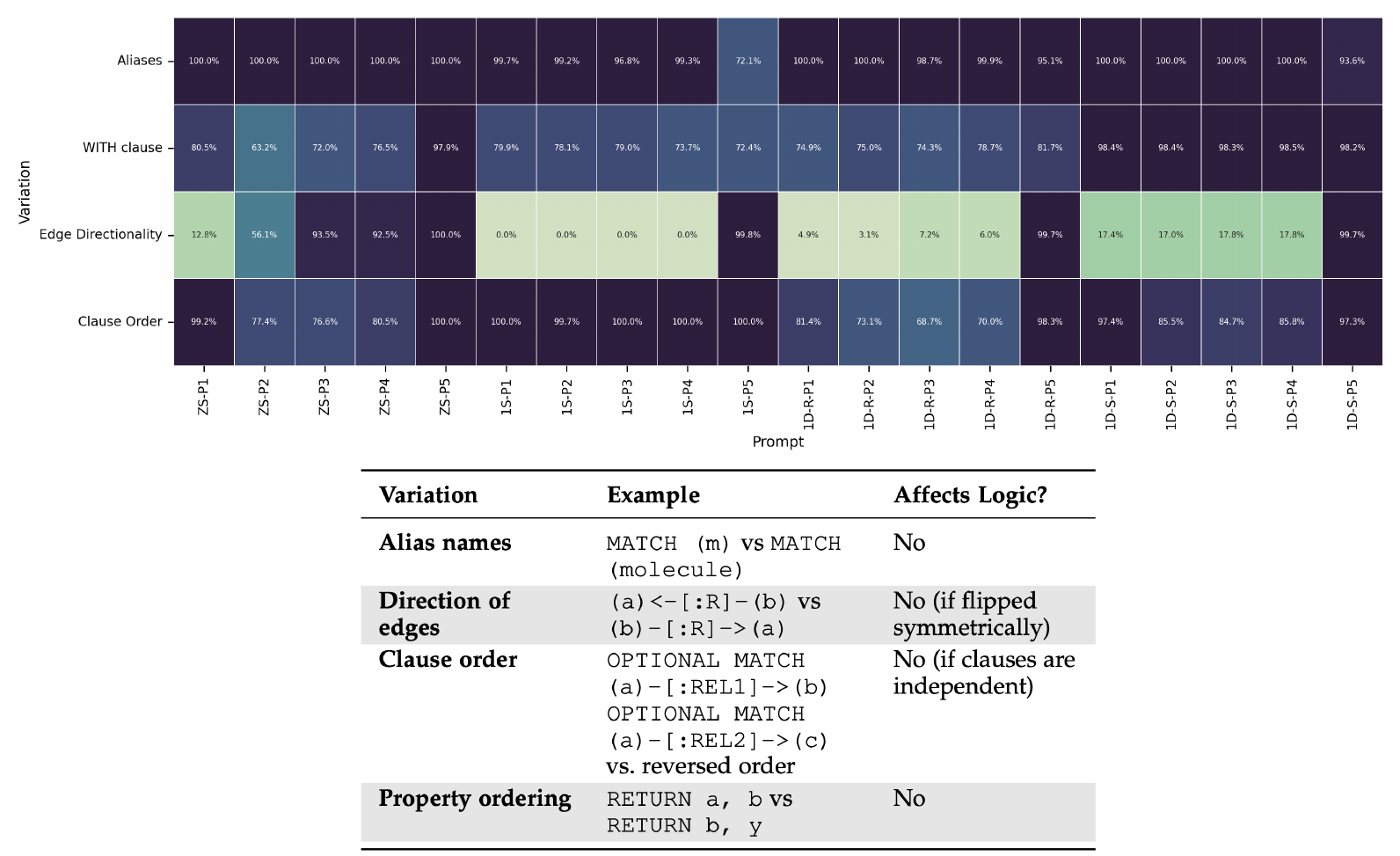}
    \caption{In the top row, we report the frequency of semantically neutral Cypher query variations, among generated queries that achieved perfect retrieval upon execution (F1=1.0), across four prompting strategies: Zero-shot (ZS), one-shot static (1S-S), one-shot random (1S-D-R) and one-shot semantic (1S-D-S). Each cell shows the percentage of queries that differ from the reference one in the corresponding category. In the bottom row, we expand more on the observed syntactic variations.}
    \label{App: tab: surface equiv}
\end{figure}

\begin{figure}[h!]
    \centering
    \includegraphics[width=1.\linewidth]{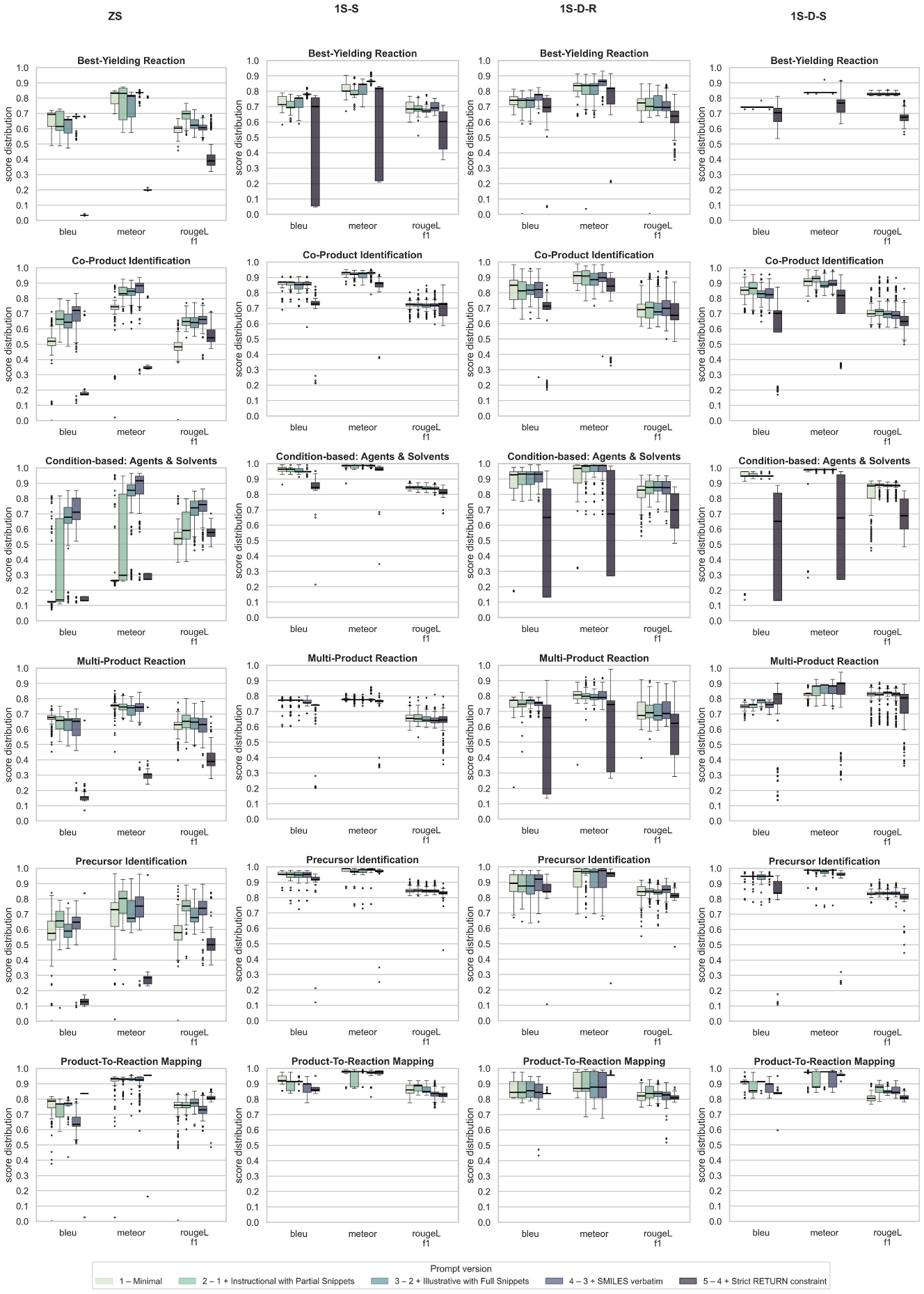}
    \caption{Comparison of generated Cypher queries to reference queries for single-step synthesis. Each subplot shows BLEU, METEOR, and ROUGE-L F1 score distributions for a specific query type under different prompt settings: Zero-shot (ZS), one-shot static (1S-S), one-shot random (1S-D-R), and one-shot semantic (1S-D-S). Within each setting, five prompt versions (color-coded) reflect increasing levels of contextual and structural guidance.}
    \label{App: fig: ST query}
\end{figure}

\begin{figure}[h!]
    \centering
    \includegraphics[width=1.\linewidth]{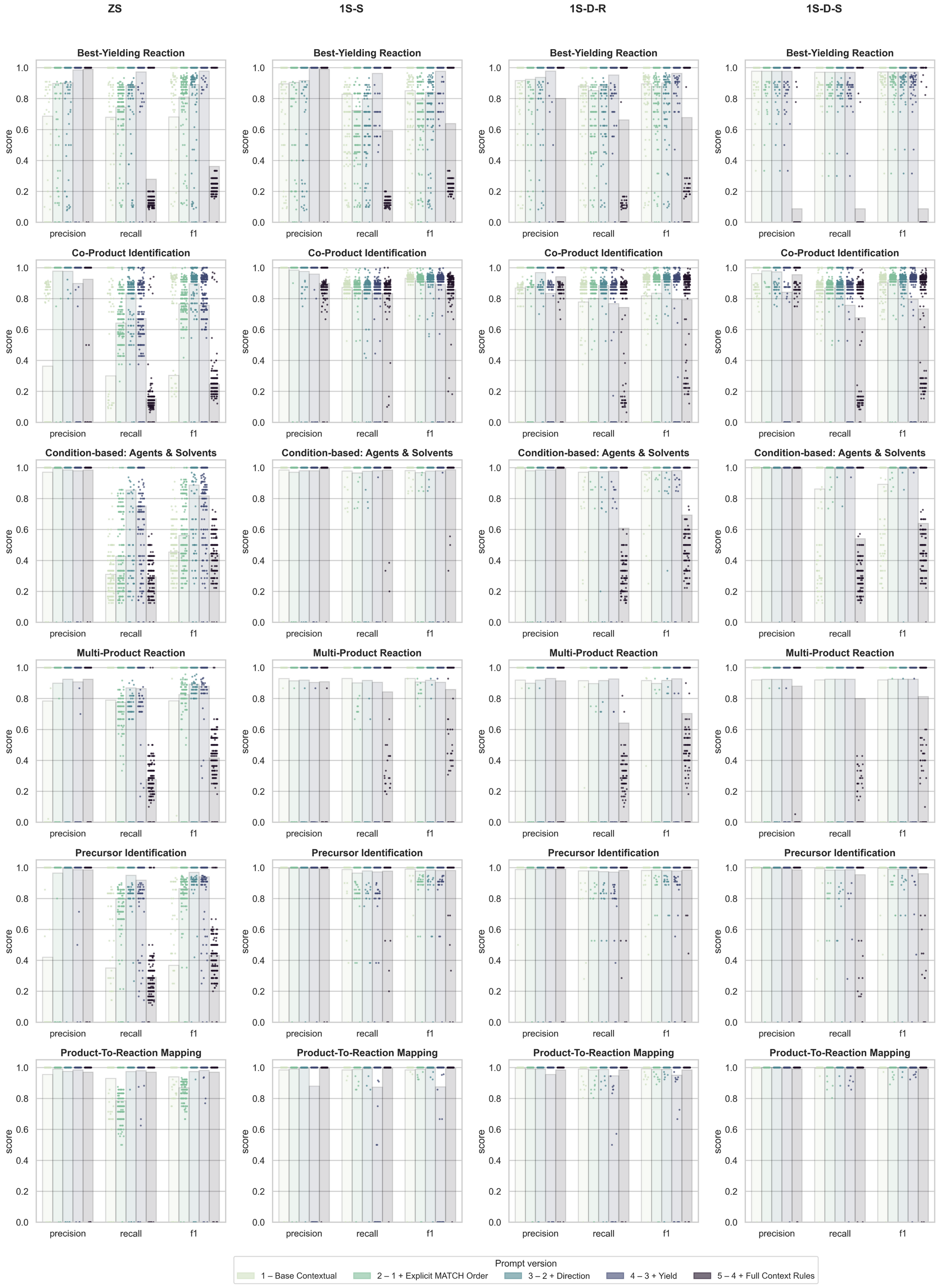}
    \caption{Single-step retrieval performance for generated Cypher queries vs ground truth. Each subplot shows precision, recall, and F1 score distributions for a specific query type under different prompt settings: Zero-shot (ZS), one-shot static (1S-S), one-shot random (1S-D-R), and one-shot semantic (1S-D-S). Within each setting, five prompt versions (color-coded) reflect increasing levels of contextual and structural guidance.}
    \label{App: fig: ST retrieval}
\end{figure}

\begin{figure}[h!]
    \centering
    \includegraphics[width=1.4\linewidth, angle=-90]{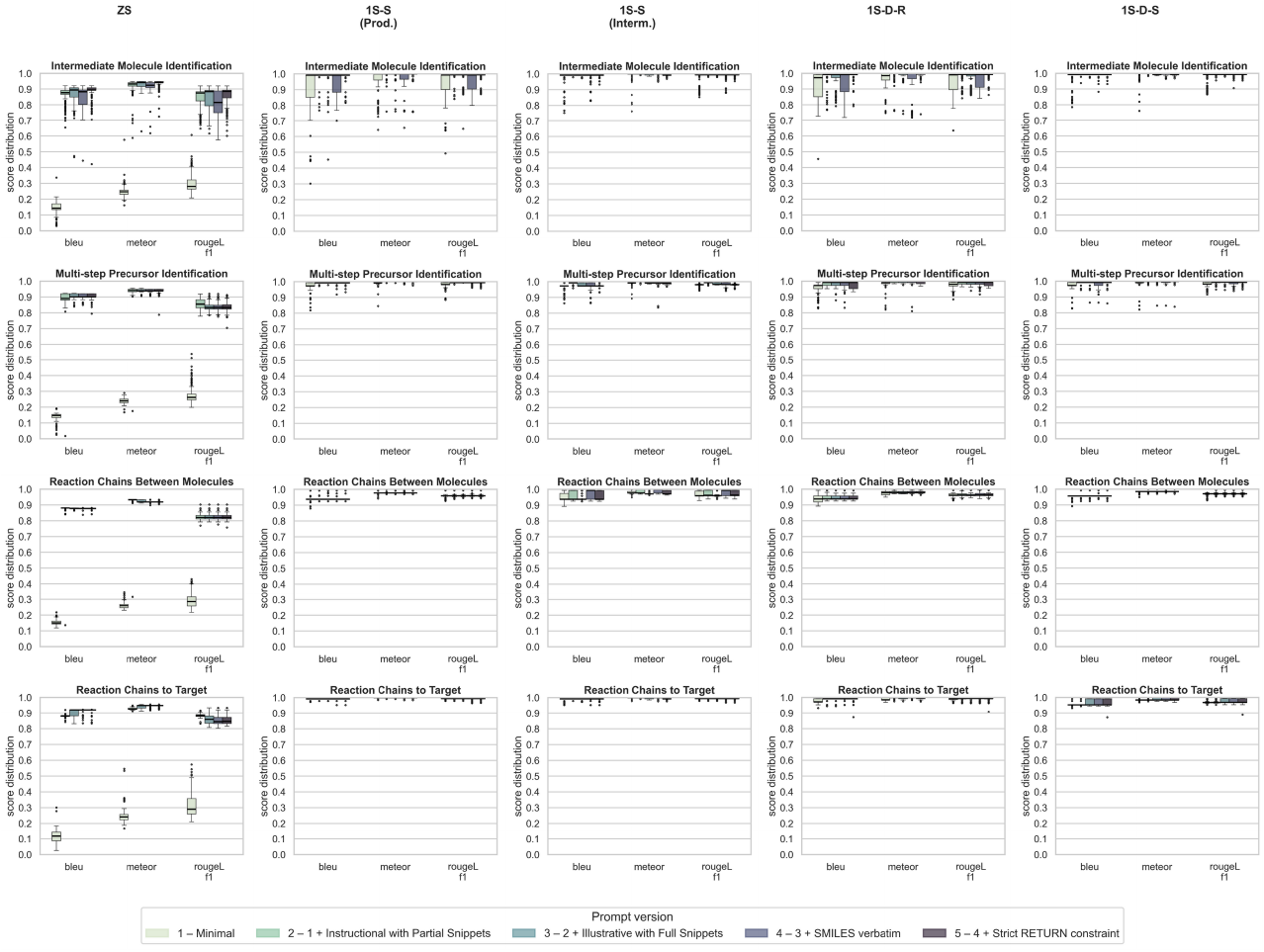}
    \caption{Comparison of generated Cypher queries to reference queries for multi-step synthesis. Each subplot shows BLEU, METEOR, and ROUGE-L F1 score distributions for a specific query type under different prompt settings: Zero-shot (ZS), one-shot static (1S-S; check \ref{App: static one-shot} for more), one-shot random (1S-D-R), and one-shot semantic (1S-D-S). Within each setting, five prompt versions (color-coded) reflect increasing levels of contextual and structural guidance.}
    \label{App: fig: MT query}
\end{figure}

\begin{figure}[h!]
    \centering
    \includegraphics[width=1.4\linewidth, angle=-90]{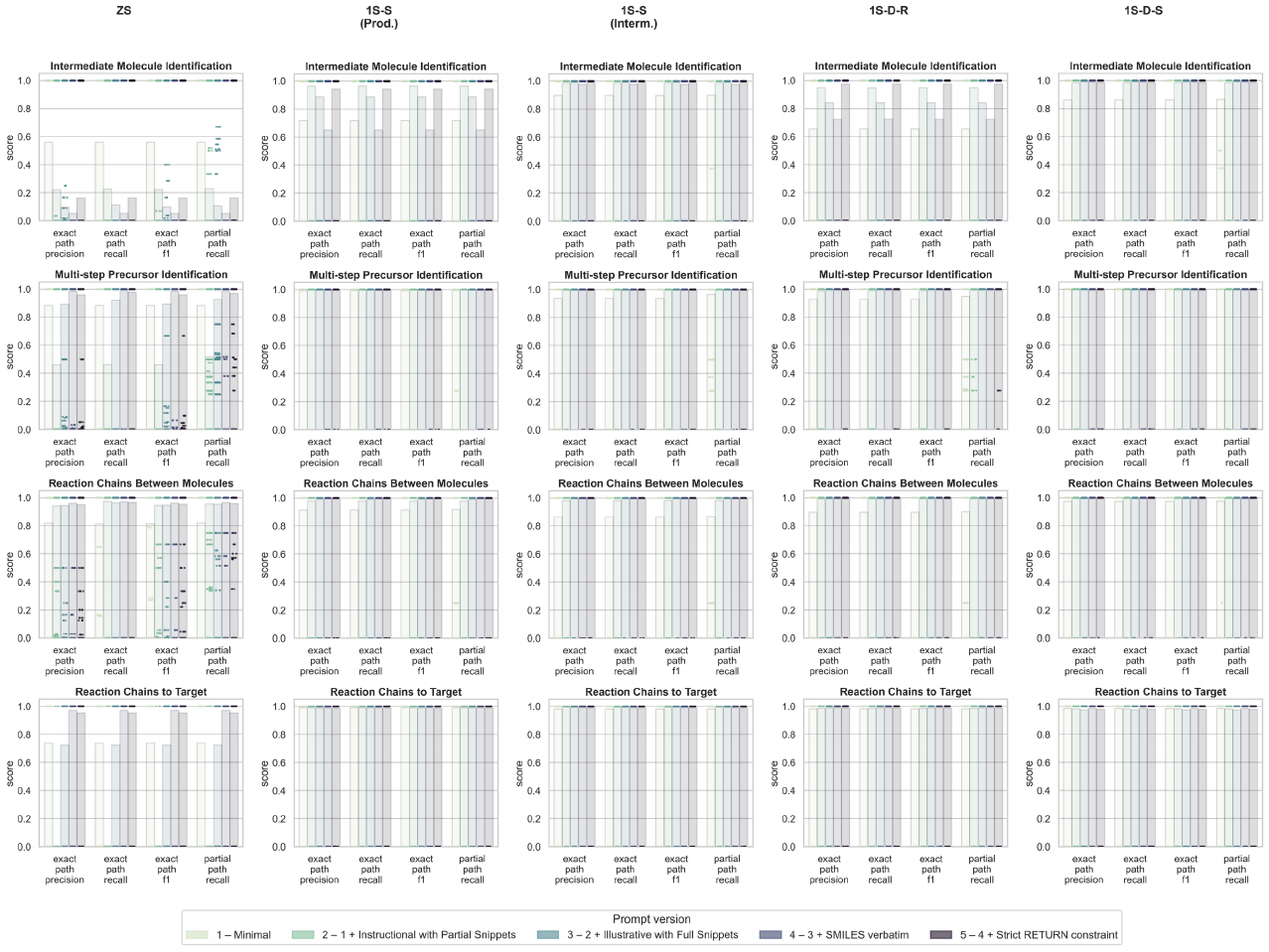}
    \caption{Multi-step retrieval performance for generated Cypher queries vs ground truth. Each subplot shows exact-path precision, recall, F1, and partial-path recall for a specific query type. We use different prompting strategies: Zero-shot (ZS), one-shot static (1S-S; check \ref{App: static one-shot} for more), one-shot random (1S-D-R), and one-shot semantic (1S-D-S). Within each setting, five prompt versions (color-coded) reflect increasing levels of contextual and structural guidance.}
    \label{App: fig: MT retrieval}
\end{figure}

Here we report results across all tasks and settings, for query generation and retrieval for single- (Figures \ref{App: tab: surface equiv}, \ref{App: fig: ST query} and \ref{App: fig: ST retrieval}) and multi-step tasks (Figures \ref{App: fig: MT query} \ref{App: fig: MT retrieval}) respectively.

\clearpage

\begin{SuppPromptWide}[s-p1]{Single-Step Synthesis Prompt  -- Base Contextual}

You are an expert in generating Cypher queries for a Neo4j knowledge graph with the following schema:

\begin{verbatim}
<schema>
{schema}
</schema>
\end{verbatim}

Your task is to generate \textbf{valid, semantically meaningful} Cypher queries based solely on user input and the schema.
\newline

\textbf{General Guidelines}
\begin{itemize}
  \item Always adhere to correct Cypher syntax.
  \item Return only the Cypher query enclosed in triple backticks --- nothing else.
\end{itemize}

\textbf{Contextual Retrieval}

 For each \texttt{Reaction} node retrieved, always include full context:
  \begin{itemize}
    \item Always use \texttt{MATCH} or \texttt{OPTIONAL MATCH} to bind variables for \texttt{reactants}, \texttt{products}, \texttt{agents}, and \texttt{solvents}.
    \item After binding, retrieve their names using \verb|collect(DISTINCT <variable>.name)| in the \texttt{RETURN} clause.
  \end{itemize}

\textbf{User Question}
\begin{verbatim}
<user_question>
{question}
</user_question>
\end{verbatim}
\end{SuppPromptWide}

\begin{SuppPromptWide}[s-p2]{Single-Step Synthesis Prompt 2 -- (1) + Explicit \texttt{MATCH} Order}
You are an expert in generating Cypher queries for a Neo4j knowledge graph with the following schema:

\begin{verbatim}
<schema>
{schema}
</schema>
\end{verbatim}

Your task is to generate \textbf{valid, semantically meaningful} Cypher queries based solely on user input and the schema.
\\

\textbf{General Guidelines}
\begin{itemize}
  \item Always adhere to correct Cypher syntax.
  \item Return only the Cypher query enclosed in triple backticks --- nothing else.
\end{itemize}
\textbf{Contextual Retrieval}

 For each \texttt{Reaction} node retrieved, always include full context:
  \begin{itemize}
    \item Always use \texttt{MATCH} or \texttt{OPTIONAL MATCH} to bind variables for \texttt{reactants}, \texttt{products}, \texttt{agents}, and \texttt{solvents}.
    \item After binding, retrieve their names using \verb|collect(DISTINCT <variable>.name)| in the \texttt{RETURN} clause.
  \end{itemize}

\textbf{User Question}
\begin{verbatim}
<user_question>
{question}
</user_question>
\end{verbatim}
\end{SuppPromptWide}

 \newpage

\begin{SuppPromptWide}[s-p3]{Single-Step Synthesis Prompt 3 -- (2) + Direction}
You are an expert in generating Cypher queries for a Neo4j knowledge graph with the following schema:

\begin{verbatim}
<schema>
{schema}
</schema>
\end{verbatim}

Your task is to generate \textbf{valid, semantically meaningful} Cypher queries based solely on user input and the schema.
\newline
\textbf{General Guidelines}
\begin{itemize}
  \item Always adhere to correct Cypher syntax.
  \item Return only the Cypher query enclosed in triple backticks --- nothing else.
\end{itemize}

\textbf{Relationship Directionality (Mandatory)} 
\newline
Ensure every relationship arrow \textbf{matches the direction shown in the schema}.
\begin{itemize}
  \item \verb|(:Molecule)-[:REACTS_IN]->(:Reaction)|
  \item \verb|(:Reaction)-[:PRODUCES]->(:Molecule)|
  \item \verb|(:Reaction)-[:USES_AGENT]->(:Molecule)|
  \item \verb|(:Reaction)-[:USES_SOLVENT]->(:Molecule)|
\end{itemize}

\textbf{Contextual Retrieval}

 For each \texttt{Reaction} node retrieved, always include full context:
  \begin{itemize}
    \item Always use \texttt{MATCH} or \texttt{OPTIONAL MATCH} to bind variables for \texttt{reactants}, \texttt{products}, \texttt{agents}, and \texttt{solvents}.
    \item After binding, retrieve their names using \verb|collect(DISTINCT <variable>.name)| in the \texttt{RETURN} clause.
  \end{itemize}

\textbf{User Question}
\begin{verbatim}
<user_question>
{question}
</user_question>
\end{verbatim}
\label{prompt:single-step-3}
\end{SuppPromptWide}

\begin{SuppPromptWide}[s-p4]{Single-Step Synthesis Prompt 4 -- (3) + Yield}\label{prompt:single-step-4}

You are an expert in generating Cypher queries for a Neo4j knowledge graph with the following schema:

\begin{verbatim}
<schema>
{schema}
</schema>
\end{verbatim}

Your task is to generate \textbf{valid, semantically meaningful} Cypher queries based solely on user input and the schema.
\newline

\textbf{General Guidelines}
\begin{itemize}
  \item Always adhere to correct Cypher syntax.
  \item Return only the Cypher query enclosed in \verb|```| --- nothing else.
\end{itemize}

\textbf{Yield Handling}

If yield is used (e.g., in filter or sort), add \verb|WHERE <rel>.yield IS NOT NULL| before any sorting or limiting.
\newline

\textbf{Relationship Directionality (Mandatory)}

Ensure every relationship arrow \textbf{matches the direction shown in the schema}:
\begin{itemize}
  \item \verb|(:Molecule)-[:REACTS_IN]->(:Reaction)|
  \item \verb|(:Reaction)-[:PRODUCES]->(:Molecule)|
  \item \verb|(:Reaction)-[:USES_AGENT]->(:Molecule)|
  \item \verb|(:Reaction)-[:USES_SOLVENT]->(:Molecule)|
\end{itemize}

\textbf{Contextual Retrieval}

For each \texttt{Reaction} node retrieved, always include full context:
\begin{itemize}
  \item Always use \texttt{MATCH} or \texttt{OPTIONAL MATCH} to bind variables for \texttt{reactants}, \texttt{products}, \texttt{agents}, and \texttt{solvents}.
  \item After binding, retrieve their names using \verb|collect(DISTINCT <variable>.name)| in the \texttt{RETURN} clause.
\end{itemize}
\textbf{User Question}
\begin{verbatim}
<user_question>
{question}
</user_question>
\end{verbatim}
\end{SuppPromptWide}

\begin{SuppPromptWide}[s-p5]{Single-Step Synthesis Prompt 5 -- (4) + Full Context Rules}

You are an expert in generating Cypher queries for a Neo4j knowledge graph with the following schema:

\begin{verbatim}
<schema>
{schema}
</schema>
\end{verbatim}

Your task is to generate \textbf{valid, semantically meaningful} Cypher queries based solely on user input and the schema.

\textbf{General Guidelines}

\begin{itemize}
  \item Always adhere to correct Cypher syntax.
  \item Return only the Cypher query enclosed in \verb|```| — nothing else.
\end{itemize}

\textbf{Yield Handling}
If yield is used (e.g., in filter or sort), add \verb|WHERE <rel>.yield IS NOT NULL| before any sorting or limiting.
\\
\textbf{Relationship Directionality (Mandatory)}

Ensure every relationship arrow \textbf{matches the direction shown in the schema}:
\begin{itemize}
  \item \verb|(:Molecule)-[:REACTS_IN]->(:Reaction)|
  \item \verb|(:Reaction)-[:PRODUCES]->(:Molecule)|
  \item \verb|(:Reaction)-[:USES_AGENT]->(:Molecule)|
  \item \verb|(:Reaction)-[:USES_SOLVENT]->(:Molecule)|
\end{itemize}

\textbf{Contextual Retrieval}

When the query involves a molecule (e.g., asking what produces it, what it reacts in, or its precursors) — follow this pattern:

\begin{enumerate}
  \item Match all reactions involving the molecule, for example: \\
        \verb|MATCH (target:Molecule {{name: "..."}})<-[:PRODUCES]-(r:Reaction)|
  \item Use \texttt{OPTIONAL MATCH} to retrieve all possible related molecules:
    \begin{itemize}
      \item \verb|(reactant:Molecule)-[:REACTS_IN]->(r:Reaction)|
      \item \verb|(r:Reaction)-[:PRODUCES]->(product:Molecule)|
      \item \verb|(r:Reaction)-[:USES_AGENT]->(agent:Molecule)|
      \item \verb|(r:Reaction)-[:USES_SOLVENT]->(solvent:Molecule)|
    \end{itemize}
  \item Return them as \verb|collect(DISTINCT …)| lists, e.g.:
\begin{verbatim}
RETURN r.id,
       collect(DISTINCT reactant.name) AS reactants,
       collect(DISTINCT product.name)  AS products,
       collect(DISTINCT agent.name)    AS agents,
       collect(DISTINCT solvent.name)  AS solvents
\end{verbatim}
\end{enumerate}
\textbf{User Question}
\begin{verbatim}
<user_question>
{question}
</user_question>
\end{verbatim}
\label{prompt:single-step-5}
\end{SuppPromptWide}

\begin{SuppPromptWide}[m-p1]{Multi-Step Synthesis Prompt 1 -- Minimal}
    You are an expert in generating Cypher queries for a Neo4j knowledge graph with the following schema:

\begin{minted}[breaklines,breakanywhere,autogobble,fontsize=\small]{text}
<schema>
{schema}
</schema>
\end{minted}

Your task is to generate \textbf{valid, semantically meaningful} Cypher queries based solely on user input and the schema.

\textbf{General Guidelines:}
\begin{itemize}
  \item Always adhere to correct Cypher syntax.
  \item Return only the Cypher query enclosed in triple backticks --- nothing else.
\end{itemize}

\textbf{Important Assumptions}
\begin{itemize}
  \item Regardless of the user's question intent (e.g., asking about precursors, agents, intermediates), you must return only the full reaction chains --- that is, the \texttt{Reaction} nodes involved in the synthesis pathway.
  \item Do not perform any reasoning or interpretation --- simply identify the sequence of \texttt{Reaction} nodes involving the mentioned molecular entities.
\end{itemize}

\textbf{Graph Constraints:}
\begin{itemize}
  \item The graph is bipartite: \texttt{Molecule} and \texttt{Reaction} nodes alternate.
  \item A synthesis path of $N$ steps has $2 \times N$ hops.
\end{itemize}

\textbf{User Question}
\begin{verbatim}
<user_question>
{question}
</user_question>
\end{verbatim}
\end{SuppPromptWide}

\begin{SuppPromptWide}[m-p2]{Multi-Step Synthesis Prompt 2 -- (1) + Query Generation Instructions with Partial Cypher Snippets}
You are an expert in generating Cypher queries for a Neo4j knowledge graph with the following schema:

\begin{minted}[breaklines,breakanywhere,autogobble,fontsize=\small]{text}
<schema>
{schema}
</schema>
\end{minted}

Your task is to generate \textbf{valid, semantically meaningful} Cypher queries based solely on user input and the schema.

\textbf{General Guidelines:}
\begin{itemize}
  \item Return only the Cypher query enclosed in triple backticks --- nothing else.
\end{itemize}

\textbf{Important Assumptions}
\begin{itemize}
  \item Regardless of the user's question intent (e.g., asking about precursors, agents, intermediates), you must return only the full reaction chains --- that is, the \texttt{Reaction} nodes involved in the synthesis pathway.
  \item Do not perform any reasoning or interpretation --- simply identify the sequence of \texttt{Reaction} nodes involving the mentioned molecular entities.
\end{itemize}

\textbf{Graph Structure and Path Lengths}
\begin{itemize}
  \item The graph is bipartite: \texttt{Molecule} and \texttt{Reaction} nodes alternate.
  \item A synthesis path of $N$ steps has $Y = 2 \times N$ hops (e.g., $2\!\to\!4$, $3\!\to\!6$, $4\!\to\!8$).
\end{itemize}

\textbf{Query Constraints}
\begin{itemize}
  \item Use a variable-length pattern with \verb![:REACTS_IN|PRODUCES*..Y]! to enable multi-step synthesis paths.
  \item Enforce \textit{exact} length with:
\begin{minted}[breaklines,autogobble,fontsize=\small]{text}
WHERE size(relationships(p)) = Y
\end{minted}
  \item Ensure bipartite alternation of nodes: \texttt{Molecule} (even), \texttt{Reaction} (odd).
  \item Return \texttt{DISTINCT reaction\_nodes} only.
\end{itemize}
\textbf{User Question}
\begin{verbatim}
<user_question>
{question}
</user_question>
\end{verbatim}
\end{SuppPromptWide}

\begin{SuppPromptWide}[m-p3]{Multi-Step Synthesis Prompt 3 -- (2) + Illustrative Query Generation Instructions with Full Cypher Snippets}

You are an expert in generating Cypher queries for a Neo4j knowledge graph with the following schema:

\begin{minted}[breaklines,breakanywhere,autogobble,fontsize=\small]{text}
<schema>
{schema}
</schema>
\end{minted}

Your task is to generate \textbf{valid, semantically meaningful} Cypher queries based solely on user input and the schema.

\textbf{General Guidelines:}
\begin{itemize}
  \item Always adhere to correct Cypher syntax.
  \item Return only the Cypher query enclosed in triple backticks --- nothing else.
\end{itemize}

\textbf{Important Assumptions}
\begin{itemize}
  \item Regardless of the user's question intent (e.g., asking about precursors, agents, intermediates), you must return only the full reaction chains --- that is, the reaction nodes involved in the synthesis pathway.
  \item Do not perform any reasoning or interpretation --- simply identify the sequence of reaction nodes involving the mentioned molecular entities.
\end{itemize}

\textbf{Graph Structure and Path Lengths}
The graph is \textbf{bipartite}:
\begin{itemize}
  \item Molecule and Reaction nodes alternate.
  \item A synthesis path of $N$ steps has path length $Y = 2 \times N$ hops (e.g., $2 \to 4$, $3 \to 6$, $4 \to 8$).
\end{itemize}

\textbf{Query Constraints}
\begin{enumerate}
  \item \textit{Match Multi-Hop Paths}\\
    Use a variable-length pattern with \verb![:REACTS_IN|PRODUCES*..Y]! to enable multi-step synthesis paths. Enforce \textit{exact} length with:
\begin{minted}[breaklines,autogobble,fontsize=\small]{text}
WHERE size(relationships(p)) = Y
\end{minted}

  \item \textit{Enforce Bipartite Alternation}:
\begin{minted}[breaklines,autogobble,fontsize=\small]{text}
WHERE all(i IN range(0, size(nodes(p)) - 1)
      WHERE (i % 2 = 0 AND 'Molecule' IN labels(nodes(p)[i])) OR
            (i % 2 = 1 AND 'Reaction' IN labels(nodes(p)[i])))
\end{minted}

  \item \textit{Return Only Reaction Nodes}:
\begin{minted}[breaklines,autogobble,fontsize=\small]{text}
WITH [x IN nodes(p) WHERE 'Reaction' IN labels(x)] AS reaction_nodes
RETURN DISTINCT reaction_nodes
\end{minted}
\end{enumerate}

\textbf{User Question}
\begin{verbatim}
<user_question>
{question}
</user_question>
\end{verbatim}
\end{SuppPromptWide}

\begin{SuppPromptWide}[m-p4]{Multi-Step Synthesis Prompt 4 -- (3) + SMILES verbatim}
You are an expert in generating Cypher queries for a Neo4j knowledge graph with the following schema:

\begin{minted}[breaklines,breakanywhere,autogobble,fontsize=\small]{text}
<schema>
{schema}
</schema>
\end{minted}

Your task is to generate \textbf{valid, semantically meaningful} Cypher queries based solely on user input and the schema.

\textbf{General Guidelines:}
\begin{itemize}
  \item Always adhere to correct Cypher syntax.
  \item Return only the Cypher query enclosed in triple backticks --- nothing else.
\end{itemize}

\textbf{Important Assumptions}
\begin{itemize}
  \item Regardless of the user's question intent (e.g., asking about precursors, agents, intermediates), you must return only the full reaction chains --- that is, the \texttt{Reaction} nodes involved in the synthesis pathway.
  \item Copy \texttt{SMILES} verbatim: character-for-character --- no changes to atoms, case, ring numbers, parentheses/brackets, or charges.
  \item Do not perform any reasoning or interpretation --- simply identify the sequence of \texttt{Reaction} nodes involving the mentioned molecular entities.
\end{itemize}

\textbf{Graph Structure and Path Lengths}
The graph is \textbf{bipartite}:
\begin{itemize}
  \item \texttt{Molecule} and \texttt{Reaction} nodes alternate.
  \item A synthesis path of $N$ steps has path length $Y = 2 \times N$ hops (e.g., $2 \to 4$, $3 \to 6$, $4 \to 8$).
\end{itemize}

\textbf{Query Constraints}
\begin{enumerate}
  \item \textit{Match Multi-Hop Paths}\\
    Use a variable-length pattern with \verb![:REACTS_IN|PRODUCES*..Y]! to enable multi-step synthesis paths. Enforce \textit{exact} length with:
\begin{minted}[breaklines,autogobble,fontsize=\small]{text}
WHERE size(relationships(p)) = Y
\end{minted}

  \item \textit{Enforce Bipartite Alternation}:
\begin{minted}[breaklines,autogobble,fontsize=\small]{text}
WHERE all(i IN range(0, size(nodes(p)) - 1)
      WHERE (i % 2 = 0 AND 'Molecule' IN labels(nodes(p)[i])) OR
            (i % 2 = 1 AND 'Reaction' IN labels(nodes(p)[i])))
\end{minted}

  \item \textit{Return Only Reaction Nodes}:
\begin{minted}[breaklines,autogobble,fontsize=\small]{text}
WITH [x IN nodes(p) WHERE 'Reaction' IN labels(x)] AS reaction_nodes
RETURN DISTINCT reaction_nodes
\end{minted}
\end{enumerate}

\textbf{User Question}
\begin{verbatim}
<user_question>
{question}
</user_question>
\end{verbatim}
\end{SuppPromptWide}

\begin{SuppPromptWide}[m-p5]{Multi-Step Synthesis Prompt 5 -- (4) + Strict \texttt{RETURN} Constraint}
You are an expert in generating Cypher queries for a Neo4j knowledge graph with the following schema:

\begin{minted}[breaklines,breakanywhere,autogobble,fontsize=\small]{text}
<schema>
{schema}
</schema>
\end{minted}

Your task is to generate \textbf{valid, semantically meaningful} Cypher queries based solely on user input and the schema.

\textbf{General Guidelines:}
\begin{itemize}
  \item Always adhere to correct Cypher syntax.
  \item Return only the Cypher query enclosed in triple backticks --- nothing else.
\end{itemize}

\textbf{Important Assumptions}
\begin{itemize}
  \item Regardless of the user's question intent (e.g., asking about precursors, agents, intermediates), you must return only the full reaction chains --- that is, the \texttt{Reaction} nodes involved in the synthesis pathway.
  \item Copy \texttt{SMILES} verbatim: character-for-character --- no changes to atoms, case, ring numbers, parentheses/brackets, or charges.
  \item Do not perform any reasoning or interpretation --- simply identify the sequence of \texttt{Reaction} nodes involving the mentioned molecular entities.
\end{itemize}

\textbf{Graph Structure and Path Lengths}
The graph is \textbf{bipartite}:
\begin{itemize}
  \item \texttt{Molecule} and \texttt{Reaction} nodes alternate.
  \item A synthesis path of $N$ steps has path length $Y = 2 \times N$ hops (e.g., $2 \to 4$, $3 \to 6$, $4 \to 8$).
\end{itemize}

\textbf{Query Constraints}
\begin{enumerate}
  \item \textit{Match Multi-Hop Paths}\\
    Use a variable-length pattern with \verb![:REACTS_IN|PRODUCES*..Y]! to enable multi-step synthesis paths. Enforce \textit{exact} length with:
\begin{minted}[breaklines,autogobble,fontsize=\small]{text}
WHERE size(relationships(p)) = Y
\end{minted}

  \item \textit{Enforce Bipartite Alternation}:
\begin{minted}[breaklines,autogobble,fontsize=\small]{text}
WHERE all(i IN range(0, size(nodes(p)) - 1)
      WHERE (i % 2 = 0 AND 'Molecule' IN labels(nodes(p)[i])) OR
            (i % 2 = 1 AND 'Reaction' IN labels(nodes(p)[i])))
\end{minted}

  \item \textit{Return Only Reaction Nodes}:
\begin{minted}[breaklines,autogobble,fontsize=\small]{text}
WITH [x IN nodes(p) WHERE 'Reaction' IN labels(x)] AS reaction_nodes
RETURN DISTINCT reaction_nodes
\end{minted}
\end{enumerate}

\textbf{Output constraint:} Only return \texttt{reaction\_nodes}. Do \emph{not} return \texttt{Molecule} nodes, \texttt{p}, \texttt{nodes(p)}, \texttt{relationships(p)}, or anything else. No extra \texttt{RETURN}s.

\textbf{User Question}
\begin{verbatim}
<user_question>
{question}
</user_question>
\end{verbatim}

\end{SuppPromptWide}

\end{document}